\documentclass{article}
\usepackage{lmodern}
\usepackage{spconf, amsmath,graphicx,hyperref}
\usepackage{enumitem}
\usepackage{amssymb}
\usepackage{booktabs}
\usepackage{multirow}
\usepackage{tabularx}
\usepackage{indentfirst}

\newcommand{\fref}[1]{Fig.~\ref{#1}}

\newcommand{\lsps}{LSPS~\cite{Woodham80} }
\newcommand{\rgbps}{RGBPS~\cite{Chakrabarti16} }
\newcommand{\ledps}{LEDPS~\cite{Queau18} }
\newcommand{\fastnfps}{FastNFPS~\cite{lichy_2022} }

\title{Near-Light Color Photometric Stereo for mono-Chromaticity non-lambertian surface}
\name{
    \begin{tabular}{c}
    Zonglin Li$^{1,\ddagger}$\thanks{$^{\ddagger}$ Equal contribution, $^{\ast}$ Corresponding author: guoheng@bupt.edu.cn}, 
    Jieji Ren$^{2,\ddagger}$, 
    Shuangfan Zhou$^{1}$, 
    Heng Guo$^{1,\ast}$, \\
    Jinnuo Zhang$^{2}$, 
    Jiang Zhou$^{2}$, 
    Boxin Shi$^{3}$, 
    Zhanyu Ma$^{1}$, 
    Guoying Gu$^{2}$
    \end{tabular}
}

\address{
    $^{1}$ Beijing University of Posts and Telecommunications, Beijing, China \\ 
    $^{2}$ Shanghai Jiao Tong University, Shanghai, China \\ 
    $^{3}$ Peking University, Beijing, China
}

\vspace{-10pt}

\begin{document}
%\ninept
%
\maketitle
\begin{abstract}
% \vspace{-1ex}

% Conventional photometric stereo relies on capturing multiple images of a static scene at different times, each under carefully controlled varying illumination from a fixed viewpoint, limiting its practicality in dynamic or time-sensitive applications. Color photometric stereo enables single-shot reconstruction, but remains constrained by distant-light assumptions and simplified reflectance models. We propose a near-light color photometric stereo method that uses neural implicit representations to jointly optimize surface geometry and reflectance from a single RGB image under colored near-light illumination, without multi-image capture or calibration. This approach is particularly suitable for compact systems like vision-based tactile sensors, enhancing the practicality of photometric stereo in tactile applications.

% Unlike conventional photometric stereo, which relies on capturing multiple images of a static scene at different timestamp, color photometric stereo enables surface reconstruction through  a single-shot image, offering advantages in applications involving snap-shot events and dynamic scenes. 
Color photometric stereo enables single-shot surface reconstruction, extending conventional photometric stereo that requires multiple images of a static scene under varying illumination to dynamic scenarios. However, most existing approaches assume ideal distant lighting and Lambertian reflectance, leaving more practical near-light conditions and non-Lambertian surfaces underexplored. To overcome this limitation, we propose a framework that leverages neural implicit representations for depth and BRDF modeling under the assumption of mono-chromaticity (uniform chromaticity and homogeneous material), which alleviates the inherent ill-posedness of color photometric stereo and allows for detailed surface recovery from just one image. Furthermore, we design a compact optical tactile sensor to validate our approach. Experiments on both synthetic and real-world datasets demonstrate that our method achieves accurate and robust surface reconstruction. To facilitate future research and ensure reproducibility, we have made our source code publicly available at \url{https://github.com/yulinjiaxue/near-light-color-ps}.

\end{abstract} 

\vspace{-1ex}


\vspace{-1.5ex}
\section{Introduction}
\vspace{-1ex}

% Recovering high-fidelity surface normals from images is a fundamental computer vision task with broad applications in robotics and industrial inspection. Photometric stereo (PS) estimates surface normals from intensity variations under different lighting conditions. 

Recovering surface normals from images is a fundamental yet challenging computer vision task with broad applications in fields such as robotics and industrial inspection. Among various techniques developed to address this problem, Photometric Stereo (PS) has proven to be an effective approach. It estimates surface normals from intensity variations under different lighting conditions.

\textbf{Conventional methods}, initially proposed by Woodham~\cite{Woodham80}, use least-squares optimization for static Lambertian surfaces and are constrained by distant illumination. Neural network-based approaches~\cite{Ikehata18,Chen20} have improved performance on non-Lambertian objects. Under near-light illumination setting, LEDPS~\cite{Queau18} handles near-light conditions through alternating optimization. FastNFPS~\cite{lichy_2022} has focused on non-Lambertian photometric stereo under uncalibrated near-field lighting. SDMUniPS~\cite{Ikehata23} generalizes the setup to the case of general unknown lighting, enabling reconstruction under more general conditions. \textit{However, conventional approaches necessitate capturing multiple images under varying illumination at different timestamps.}

% Since objects are required to be static during the capture, limiting their applications in dynamic scenarios.

% These methods rely on ground truth normals, which are difficult to obtain. Moreover, traditional approaches require both static scenes and multiple illuminations, which limits dynamic reconstruction and imposes strong acquisition constraints, thereby hindering their applicability in real-world scenarios.

\begin{figure}[t] \centering
    \includegraphics[width=0.48\textwidth]{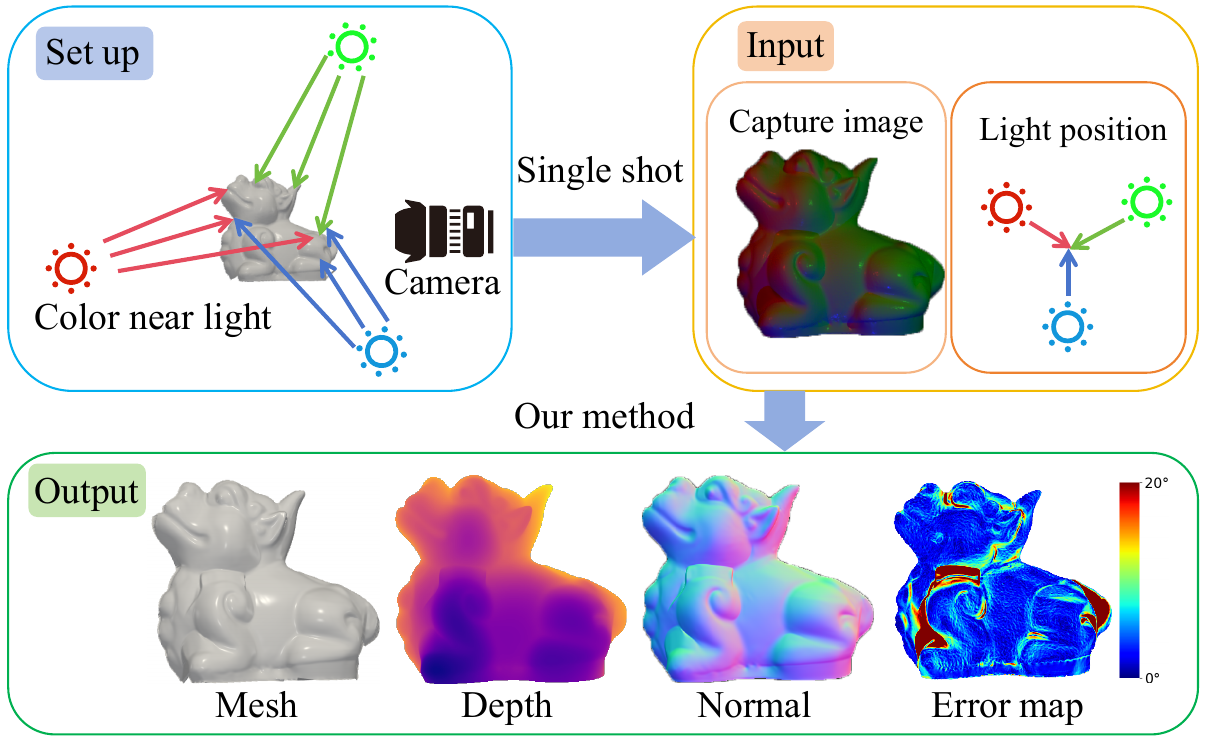}
    \vspace{-5ex}
    \caption{Our method takes a single captured RGB image and the near-light source position as input, and outputs accurate surface normal, depth map and mesh.
    } \label{fig:teaser}
    \vspace{-15pt}
\end{figure}

\begin{figure*}[t] \centering
    \includegraphics[width=\textwidth]{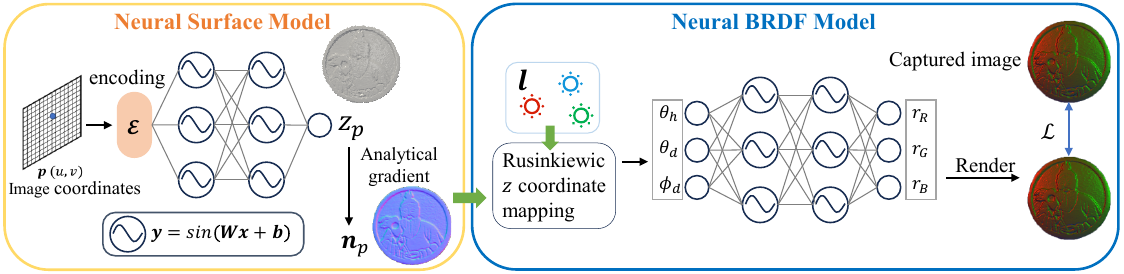}
    \vspace{-5ex}
    \caption{Our method has a neural surface model and a neural BRDF model. The neural surface model estimates depth and normal from image coordinates, the BRDF model predicts reflectance using the angle between surface normal, light and view directions. The system is optimized via photometric consistency loss between the rendered and captured images.} 
    \vspace{-3ex}
    \label{fig:pipeline}
\end{figure*}

\textbf{Color photometric stereo (CPS) } offers a single-shot alternative through spectral multiplexing, thereby enabling snapshot surface normal estimation. However, CPS is inherently ill-posed at the pixel level. This arises because, in CPS with \(k\) spectral light sources, each isolated pixel provides \(k\)  observations, but each spectral band has its own distinct spectral reflectance, the total number of unknowns becomes \(k + 2\) (\(k\) spectral reflectances plus two surface normal parameters). To transform this ill-posed problem into a well-posed one, additional priors or constraints are typically required to regularize the solution. Anderson \textit{et al.}~\cite{Anderson11a} and Y. Ju \textit{et al.}~\cite{Ju20a} employed prior information for simplification. Methods~\cite{Chakrabarti16,Ozawa18} simplified the problem by assuming the surface spectral reflectance type to be gray or mono-chromatic with uniform albedo. Guo \textit{et al.}~\cite{Guo21} leveraged multiple pixels sharing the same chromaticity \(\mathbf{v}\) to jointly constrain the solution under the assumption of mono-chromaticity. J. Lv \textit{et al.}~\cite{Lv23} addressed the non-Lambertian spectral BRDF under distant light by introducing a strong model assumption of Spectral Reflectance Decomposition, decomposing the BRDF into two simpler components. Physics-Free~\cite{ikehata2025physicsfree} achieves high-quality reconstruction of non-Lambertian surfaces even under uncalibrated distant illumination. The near-light photometric stereo method of Chen~\textit{et al.}~\cite{chen20b} is only applicable to human faces under the Lambertian reflectance assumption. \textit{The inherent ill-posedness of CPS is further exacerbated in practical near-light illumination setting and non-Lambertian reflectances.}

% To address the ill-posedness of CPS under practical near-light illumination and non-Lambertian reflectance, we consider imposing mono-chromaticity prior, which assumes that the surface is uniform chromaticity and homogeneous material. This constraint has been shown to effectively regularize the reconstruction problem under such challenging conditions. Crucially, this assumption is not merely a modeling convenience; it is naturally satisfied in several real-world sensing systems. In particular, high-resolution vision-based tactile sensors such as GelSight~\cite{Yuan17} employ a homogeneous elastomer layer as the contact interface and operate under fixed, near-field illumination in a single-shot configuration—precisely matching the conditions under which the homogeneity prior holds. This makes them an ideal application domain for CPS under this constraint. Hence, developing a photometric stereo method tailored to this constrained yet practically relevant setting offers both theoretical insight and real-world utility.

To address the ill-posedness of CPS under practical near-light illumination and non-Lambertian reflectance, we consider imposing mono-chromaticity prior, which assumes that the surface is uniform chromaticity and homogeneous material. This constraint has been shown to effectively regularize the reconstruction problem under such challenging conditions. Crucially, this assumption is not merely a modeling convenience—it is naturally satisfied in real-world sensing systems such as GelSight~\cite{Yuan17}, which uses a homogeneous elastomer interface under fixed, near-field, single-shot illumination, precisely matching the mono-chromaticity prior. This makes it an ideal application for CPS under this constraint, and motivates the development of tailored photometric stereo methods with both theoretical and practical value.

Under the assumption of mono-chromaticity, we propose a neural implicit modeling framework for near-light CPS and design a practical capture setup, as shown in \fref{fig:teaser}. We model the surface depth using the differentiable neural representation, which enables the simultaneous optimization of both depth and normal, thereby reducing the difficulty of surface shape estimation. Simultaneously, under the homogeneous material constraint, the surface reflectance exhibits a unified BRDF structure across the entire surface. This constraint allows the BRDF to be effectively represented using a single implicit neural network. Further incorporating the uniform chromaticity condition, the solution of non-Lambertian spectral BRDF becomes tractable. This approach facilitates accurate normal estimation and robust geometry reconstruction under near-light conditions. We also designed a compact near-light visuo-tactile sensor (NearLightTactile) sensor for qualitative evaluation of our method's performance in real-world scenarios.

% High-resolution tactile sensors like GelSight~\cite{Yuan17} use cameras to capture deformations of a soft gel layer under LEDs, reconstructing contact geometry through heuristic or model-based methods. This sensor setup aligns closely with the requirements of near-light color photometric stereo. 

% To validate the practical utility of our method, real-world evaluation is necessary. We note that high-resolution tactile sensors such as GelSight~\cite{Yuan17} operate in a near-field snapshot configuration, which closely matches the operational conditions required by our approach. We identify this as a promising application direction for color photometric stereo. Consequently, we designed a compact near-light visuo-tactile sensor (NearLightTactile) sensor for qualitative evaluation of our method's performance in real-world scenarios.

Our contributions can be summarized as follows:
% Additionally, to validate the practical applicability of our method, we developed a real-world near-light visuo-tactile sensor capable of capturing single-shot data under compact near-light illumination. The performance of the algorithm was qualitatively evaluated using real-world captured data. 

% \begin{itemize}
% {\vspace{-2ex}}
%     \item We transform the geometric ill-posedness under near-light conditions into a well-posed problem through implicit neural depth representation, enabling stable inference of surface normals and 3D structure.
%     {\vspace{-1.5ex}}
%     \item We convert the ill-posedness of non-Lambertian spectral reflectance into a well-posed inference task via a mono-chromaticity constraint and implicit BRDF modeling, allowing robust material and reflectance recovery under color photometric stereo.
% \end{itemize} 
{\vspace{-1.5ex}}

\begin{itemize}
    \item We propose the first near-light CPS method, allowing accurate normal recovery for non-Lambertian surface under practical near light setting;
    \vspace{-1.5ex}
    \item We propose to use a neural surface model to jointly recover depth and normal, and a neural BRDF model for mono-chromaticity BRDF, formulating the near-light CPS to be well-posed;
    \vspace{-1.5ex}
    \item We design a tactile sensor \textbf{NearLightTactile}, validating our method in tactile-based shape reconstruction.
\end{itemize} 
{\vspace{-2.5ex}}

\vspace{-1ex}
\section{Method}
\vspace{-1ex}
% \label{sec:majhead}
% Our self-supervised framework aims to recover high-fidelity surface normals from a single RGB image captured under known near-light illumination, eliminating the need for ground-truth normal data or complex calibration procedures. The core of our approach is a differentiable rendering pipeline that jointly optimizes a neural surface representation and a neural BRDF model by enforcing photometric consistency between the input and rendered images. The overall architecture is depicted in Fig. \ref{fig:figure1}.

Our self-supervised framework recovers high-fidelity surface normals from a single RGB image under known near-light illumination, without ground-truth data or complex calibration. We jointly optimize a neural surface and a neural BRDF model through  photometric consistency loss between the rendered and input, without requiring ground-truth normals. The overall architecture is shown in Fig. \ref{fig:pipeline}.

\vspace{-9pt}
\subsection{Neural Surface Model}
\vspace{-1.5ex}
Instead of directly predicting surface normals, we estimate per-pixel depth values and subsequently derive normals analytically, thereby resolving the ill-posedness arising from geometric coupling in near-light CPS. 

Given a pixel coordinate $\mathbf{p}=(u,v)^\top \in \mathbb{R}^2$, we first embed it using a multi-resolution hash encoding $\mathcal{E}(\cdot)$ \cite{Muller22}, which preserves high-frequency spatial details. The encoded feature is then mapped to a scalar depth value $z$ via a sinusoidal MLP (SIREN~\cite{Sitzmann20}):

\vspace{-1ex}
\begin{equation}
z = f^{\text{SURFACE}}\!\big(\mathcal{E}(\mathbf{p});\,\theta_\text{s}\big),
\label{eq:depth}
\vspace{-1ex}
\end{equation}
where $\theta_\text{s}$ denotes the learnable parameters. The surface normal $\mathbf{n}$ is subsequently derived from the spatial gradient of the depth map $\nabla z = (\partial z/\partial u,\,\partial z/\partial v)^\top$ using the perspective camera model~\cite{Queau18survey}:

\vspace{-1ex}
\begin{equation}
\mathbf{n} = \eta\left(
\begin{bmatrix}
f\,\nabla z \\
-\,z - \nabla z^\top \mathbf{p}
\end{bmatrix}
\right),
\label{eq:normal}
\vspace{-1ex}
\end{equation}
where $\eta(\cdot)$ denotes $\ell_2$ normalization.
%%%%%%%%%%
\vspace{-9pt}
\subsection{Neural BRDF Model}
\vspace{-1.5ex}
With geometry (depth and normals) recovered from the first module, we further model spatially-varying reflectance following~\cite{Guo23}. For a surface point $\mathbf{x}$ with normal $\mathbf{n}$ and a near-light source at position $\mathbf{l}$, the normalized incident light direction is $ \mathbf{q} = \eta(\mathbf{l}-\mathbf{x})$.
% \vspace{-1ex}
% \begin{equation}
% \mathbf{L} = \eta(\mathbf{p}_{\mathrm{light}}-\mathbf{x}).
% \label{eq:light}
% \end{equation}
The pair $(\mathbf{q}, \mathbf{n})$ is then transformed into the Rusinkiewicz parameterization $(\theta_h,\theta_d,\phi_d)$ \cite{Rusinkiewicz98}. We learn the mapping from $(\theta_h,\theta_d,\phi_d)$ to the reflectance $\mathbf{r}$ using an implicit neural BRDF model:
% \vspace{-1ex}
\begin{equation}
\vspace{-1ex}
\mathbf{r} = f^{\text{BRDF}}\!\big((\theta_h,\theta_d,\phi_d);\,\theta_\text{r}\big),
\label{eq:brdf}
% \vspace{-1ex}
\end{equation}
where $\theta_\text{r}$ denotes the learnable parameters. To capture wavelength-dependent reflectance under colored LED illumination, we employ independent MLP branches for the RGB channels and concatenate their outputs as $\mathbf{r} = [r_R,\,r_G,\,r_B]$.
% \vspace{-1ex}
% \begin{equation}
% \mathbf{r} = [r_R,\,r_G,\,r_B].
% \label{eq:rgb}
% \end{equation}

%%%%%%%%%%

\vspace{-9pt}
\subsection{Differentiable Rendering and Optimization}
\vspace{-1.5ex}

The learning process is guided by a reconstruction loss that compares the input image with a rendered image synthesized from the estimated geometry and reflectance. The RGB value $\hat{I}_{u,v}$ at pixel $(u,v)$ is computed using a near-light image formation model:
\vspace{-0.5ex}
\begin{equation}
\hat{I}_{u,v} = \mathbf{r} \cdot \frac{\varphi}{\|\mathbf{l} - \mathbf{x}\|^2} 
\max(\mathbf{q}^\top \mathbf{n}, 0),
\label{eq:render}
\vspace{-1ex}
\end{equation}
where $\varphi$ denotes the light intensity, set to $1$ in our implementation. 
This image formation model accounts for both light attenuation
($1/\|\mathbf{l} - \mathbf{x}\|^2$) and cosine falloff ($\max(\mathbf{q}^\top \mathbf{n}, 0)$).

The network is trained end-to-end by minimizing an $\ell_1$ reconstruction loss:
\vspace{-1ex}
\begin{equation}
\mathcal{L} = \sum_{u,v} \| I_{u,v} - \hat{I}_{u,v} \|_1.
\label{eq:loss}
\vspace{-1.5ex}
\end{equation}

% Gradients are backpropagated through the differentiable renderer to jointly optimize the neural surface parameters ($\theta_s$) and neural BRDF parameters ($\theta_r$), ensuring both geometric and photometric consistency.

% \vspace{-2ex}
\section{Experiments}
\vspace{-1ex}
% \label{sec:print}

\subsection{Experimental Setup}
\vspace{-1.5ex}
% \label{ssec:subhead}

% For non-Lambertian objects, we use Blender to render 5 distinct objects with diverse geometries (flat, curved, rugged) and reflectance properties (Lambertian, metallic, ceramic). 

% To simulate real-world imaging conditions, it is necessary to configure parameters in Blender that are consistent with actual settings. The relevant parameters are as follows: The object is positioned 35 mm away from the camera. The camera sensor size is set to 2.3 mm × 1.44 mm, with a focal length of 2.8 mm. The capture resolution of the camera is 640 × 480.A point light source is used, positioned 11 mm away from the camera in the longitudinal direction and 21.5 mm in the lateral direction. The light intensity should be within a reasonable range to avoid overexposure or underexposure.

% For each object, we generate a single RGB image under three near-light RGB LEDs, along with the corresponding ground-truth normal map and object mask. We compare our method against several classical: LSPS, LEDPS, RGB-PS.We use the standard Mean Angular Error (MAE) in degrees to measure the accuracy of the predicted surface normals against the ground truth.

Due to the lack of publicly available CPS datasets captured under near-light conditions, we generate a synthetic dataset of five objects using Blender v3.6\footnote{https://www.blender.org/}, designed to cover variations in color, reflectance, and geometry. Our dataset comprises both white and colored objects to examine robustness against intrinsic color; Lambertian and non-Lambertian materials (e.g., metal, ceramic) to test reflectance modeling; and objects ranging from flat to curved and rugged surfaces to evaluate reconstruction across different geometric complexities. All objects are rendered with Blender’s principled BSDF shader. The camera is positioned 35 mm away from the object, with a sensor size of $2.3 \times 1.44$ mm, focal length of 2.8 mm, and image resolution of $640 \times 480$. A point light source is placed 11 mm away from the camera longitudinally and 21.5 mm laterally. Three colored LEDs (red, green, blue) are simulated to provide near-light illumination, and their intensity is tuned to avoid over-exposure. Rendering is performed in Blender Cycles using a linear color space to preserve radiometric consistency.

Since most CPS methods require constraints from four or more light sources, we ultimately opted to compare our method with LSPS \cite{Woodham80}, LEDPS \cite{Queau18}, RGB-PS \cite{Chakrabarti16} and FastNFPS \cite{lichy_2022}, using their default implementations where available. Performance is evaluated with the standard Mean Angular Error (MAE) in degrees between predicted and ground-truth normals.

\begin{figure}[t] \centering
    \includegraphics[width=0.48\textwidth]{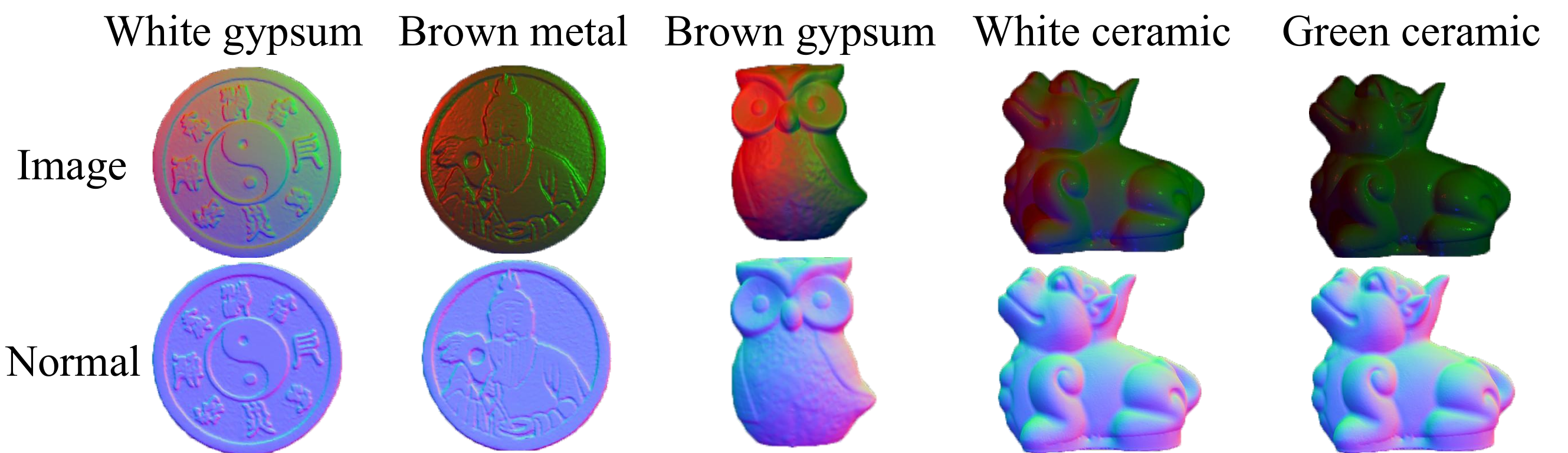}
    \vspace{-5ex}
    \caption{
    % The five datasets we used for evaluation cover different reflection types, surface types, and varying colors.
    Synthetic dataset covering varying geometric details and reflection types.
    } \label{fig:data}
    \vspace{-3ex}
\end{figure}

\begin{table}[t]
    \centering
    \caption{Quantitative comparison of surface normals}
    \vspace{1ex}
    \label{tab:quantitative}
    \renewcommand{\arraystretch}{1.5}
    \resizebox{0.45\textwidth}{!}{
    \fontsize{22}{18}\selectfont
    \begin{tabular}{cccccc}
        \toprule
        \multirow{2}{*}{Object} & 
        \multicolumn{5}{c}{Mean Angular Error($^\circ$) $\downarrow$} \\
        \cmidrule(lr){2-6}
         & Ours & \lsps & \ledps & \rgbps & \fastnfps \\
        \midrule
        White gypsum & \textbf{2.54} & 17.98 & 3.55 & 17.91 & 25.64 \\
        Brown metal & \textbf{4.74} & 24.67 & 13.11 & 20.41 & 21.19 \\
        Brown gypsum & \textbf{2.43} & 13.84 & 6.41 & 13.98 & 12.04 \\
        White ceramic & \textbf{6.38} & 17.96 & 11.89 & 18.50 & 18.48 \\
        Green ceramic & \textbf{5.57} & 17.82 & 11.77 & 18.14 & 18.10 \\
        \bottomrule
    \end{tabular}
    }
    \vspace{-3ex}
\end{table}

\begin{figure}[t] \centering
    \includegraphics[width=0.45\textwidth]{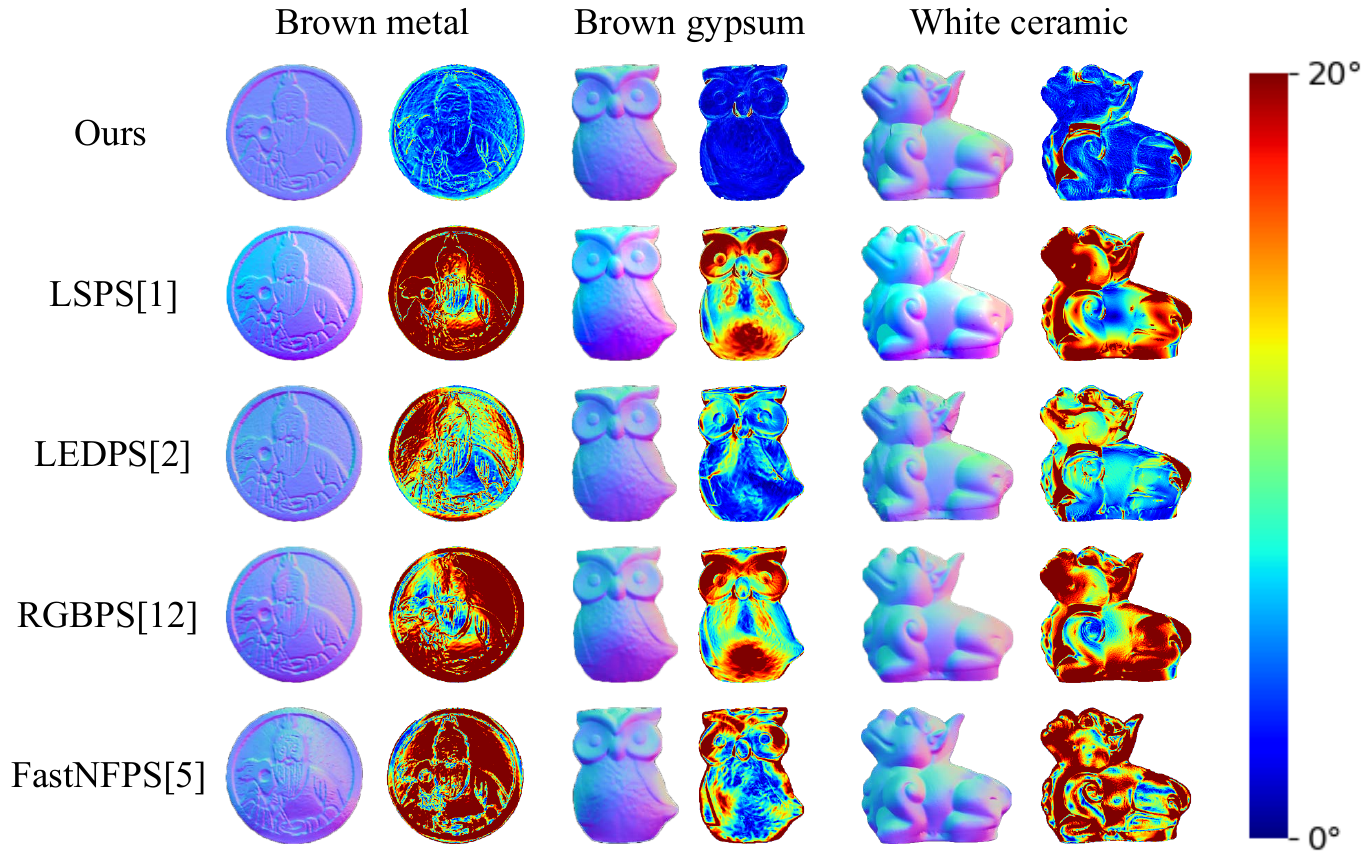}
    \vspace{-2.5ex}
    \caption{
    % Comparative evaluation results. Bar plot shows the Mean Angular Error (MAE°) for our method and baselines. Corresponding normal map visualizations demonstrate our method's lower error and higher fidelity in reconstructing surface details.
     Quantitative comparison of surface normal.
    % with normal maps confirming higher reconstruction fidelity and detail accuracy.
    } \label{fig:qualitative result}
    \vspace{-5pt}
\end{figure}

\vspace{-9pt}
\subsection{Experiment Results and Analysis}
\vspace{-1.5ex}
Quantitative results are summarized in Table \ref{tab:quantitative} and three qualitative results are illustrated in \fref{fig:qualitative result}. Our method consistently produces reconstructions that are visually closest to the ground truth, accurately preserving fine geometric details. The distant-light methods \lsps and \rgbps are not suitable for near-light conditions, resulting in poor reconstruction performance. \ledps, on the other hand, performs poorly when dealing with non-Lambertian objects. \fastnfps method does not perform well on our three-light, sub-millimeter-scale dataset. We attribute this degradation to two main factors: (1) its lighting calibration network is unstable under near-light conditions with very few sources, resulting in inaccurate light estimation; (2) its training data does not cover such lighting-to-object scale ratios, which restricts generalization. This underscores the significant challenges inherent in near-light reconstruction for sub-millimeter-scale scenarios.

% The quantitative results on the synthetic dataset indicate that our method indirectly solves the geometric decoupling of surface normals through implicit neural networks, while decoupling non Lambertian spectral BRDF using mono-Chromaticity constraints and implicit expression of BRDF. By decoupling these two aspects, the ill-posed problem of color photometric stereo on near light sources has been solved, and relatively good reconstruction accuracy has been demonstrated.
The quantitative results on the synthetic dataset demonstrate that our method reformulates the originally ill-posed problem of CPS under near-light illumination into a well-posed one. This is achieved by implicitly disentangling geometry through neural representations while decoupling non-Lambertian spectral BRDF using mono-chromaticity constraints and an implicit BRDF formulation. Through this two-fold decoupling, we show improved reconstruction accuracy under challenging near-light conditions.

\vspace{-9pt}
\subsection{Ablation Study}
\vspace{-1.5ex}
% \label{ssec:subhead}

\begin{figure}[t] \centering
    \includegraphics[width=0.4\textwidth]{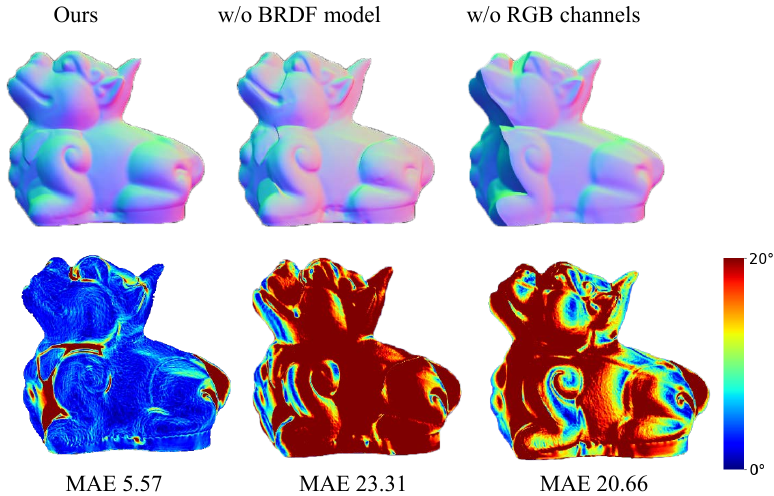}
    \vspace{-2.5ex}
    \caption{
    % Quantitative and qualitative results of ablation studies, it demonstrate that our module design achieves superior performance.
    Ablation Study of our method.
    } \label{fig:ablation}
    \vspace{-12pt}
\end{figure}

% We conduct an ablation study on the neural BRDF module to validate our design choices, with results summarized in Fig. \ref{fig:figure3}

% w/o Reflectance Module: Removing the neural BRDF and using a least-squares estimated albedo 
% causes a significant performance drop, confirming the necessity of explicit non-Lambertian modeling.

% w/o Independent RGB Outputs: Using a single output for all color channels instead of three independent ones also degrades performance. This validates that modeling wavelength-dependent reflectance is crucial under colored lighting.

We perform two ablation studies on the neural BRDF model to validate our design choices, with results shown in Fig.~\ref{fig:ablation}. (1) w/o BRDF model: Replacing the neural BRDF with a least-squares estimated albedo leads to a substantial performance drop, highlighting the importance of explicitly modeling non-Lambertian reflectance. (2) w/o RGB channels: Predicting a single output for all color channels instead of three independent ones also degrades performance, demonstrating that wavelength-dependent reflectance modeling is essential under colored lighting.

\vspace{-9pt}
\subsection{Real-World Validation}
\vspace{-1.5ex}
% \label{ssec:subhead}

\begin{figure}[t] \centering
    \includegraphics[width=0.48\textwidth]{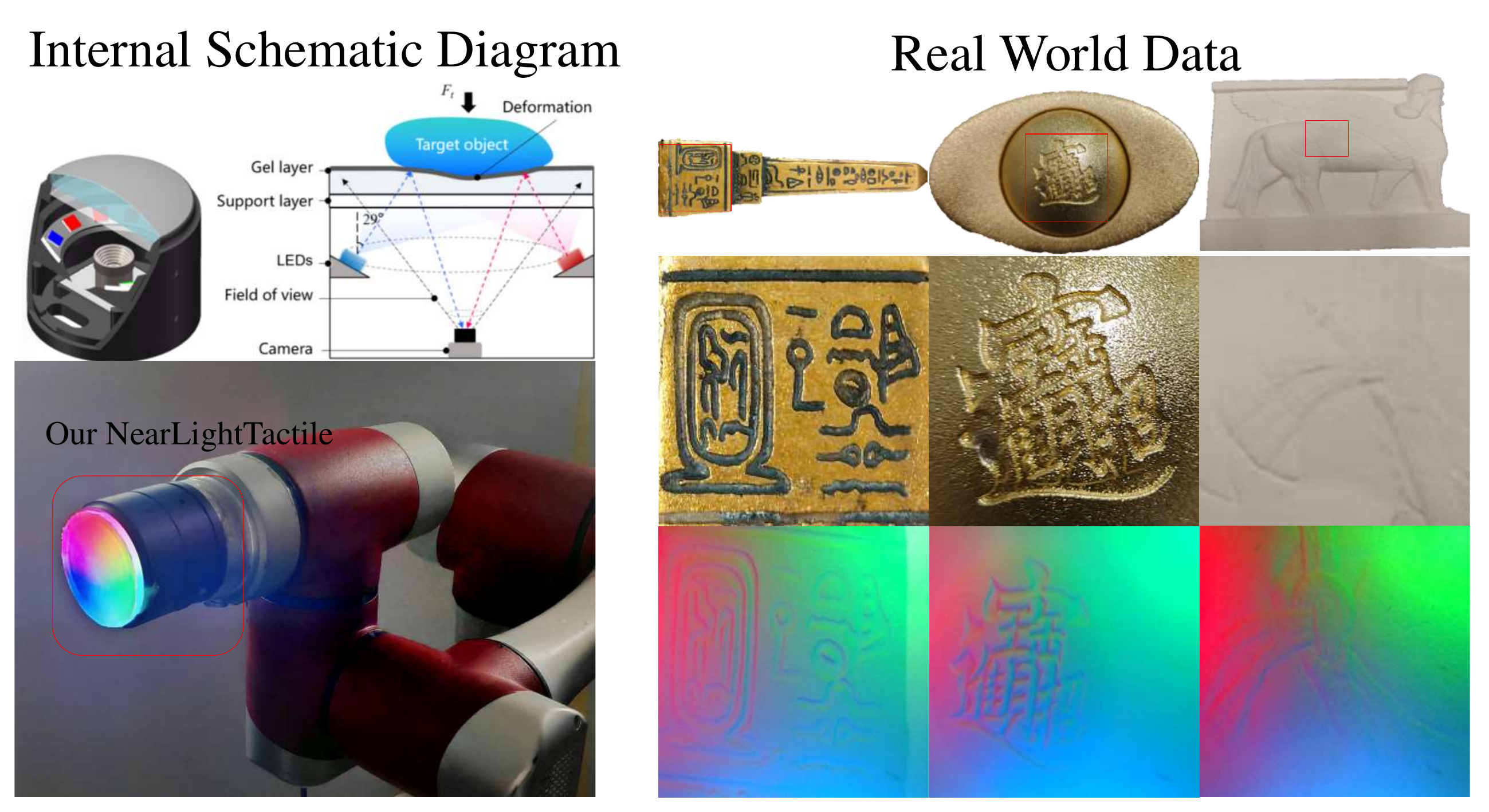}
    % \vspace{-5ex}
    \caption{
    % Real-world validation of the vision-based tactile sensor. The operational principle and a physical photograph of our custom-designed sensor are shown on the left side of the figure, while sample images of objects captured in a real-world setting are presented on the right.
    Real-world validation uses our custom vision-based tactile sensor, shown alongside sample captured images.
    } \label{fig:real}
    \vspace{-12pt}
\end{figure}

% High-resolution tactile sensors like GelSight~\cite{Yuan17} use cameras to capture deformations of a soft gel layer under LEDs, reconstructing contact geometry through heuristic or model-based methods. This sensor setup aligns closely with the requirements of near-light color photometric stereo. We identify this as a promising application direction for color photometric stereo. Consequently, we designed a \textit{NearLightTactile} sensor for qualitative evaluation of our method's performance in real-world scenarios
As detailed in Fig. \ref{fig:real}, this sensor provides a challenging real-world testbed characterized by near-light illumination, single imaging and sub-millimeter geometric features.

\vspace{-3ex}
\subsubsection{Robustness against Crosstalk}
\vspace{-1ex}

Crosstalk~\cite{Ahmad19} poses a major challenge for CPS in compact systems , due to spectral overlap in RGB filters and pixel-level cross-coupling in miniaturized sensors. This violates the channel independence assumption in RGB-PS and causes significant artifacts.

To overcome Crosstalk, we propose a lightweight data-driven Crosstalk Correction Module (CCM) that models crosstalk as a pixel-wise mapping $\mathcal{T}$ from $\mathbf{I}{\text{obs}}$ to $\mathbf{I}{\text{ideal}}$. Using baseline images ${\mathbf{I}_R, \mathbf{I}_G, \mathbf{I}_B}$ captured under separate monochromatic LEDs, we use a compact MLP to learn $\mathcal{T}$, which serves as a pre-processing step to restore photometric consistency before normal estimation.

\vspace{-3ex}
\subsubsection{Qualitative Results on Real-World Data}
\vspace{-1ex}
% \label{sssec:subsubhead}

\begin{figure}[t] \centering
    \includegraphics[width=0.48\textwidth]{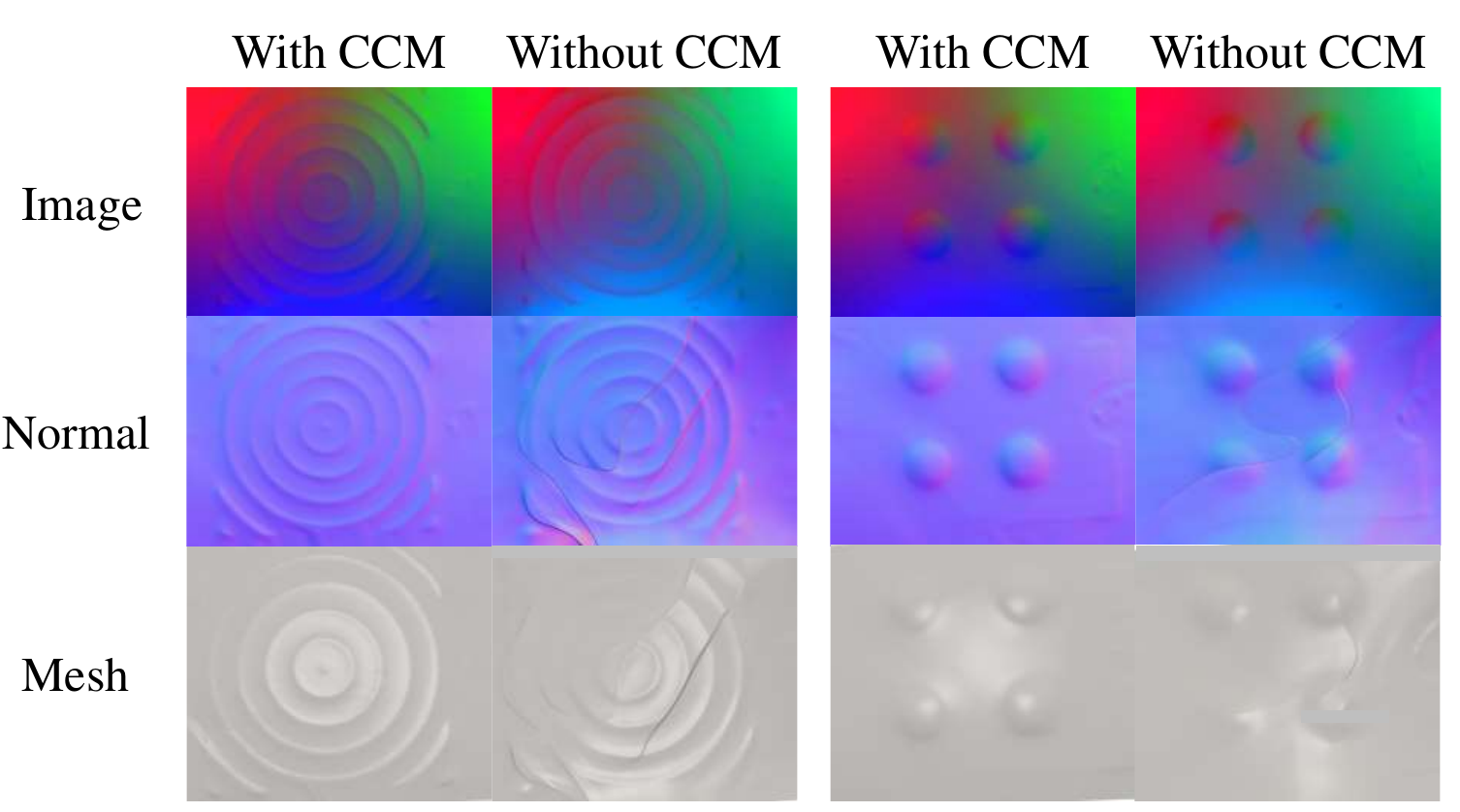}
    \vspace{-5ex}
    \caption{
    % Results on real-world captured data. This figure presents the results from three representative real-world objects with distinct shapes, captured using our sensor. For each object, the results include the crosstalk-corrected image, the surface normal map reconstructed by our algorithm, and the corresponding 3D mesh.
    % Real-world results on two distinct objects, showing images before and after crosstalk correction, along with their corresponding reconstructed normal maps and 3D meshes.
Results after and before using the CCM.
    } \label{fig:real result}
    \vspace{-15pt}
\end{figure}

We evaluate our complete system (sensor + CCM + method) by pressing objects with fine, sub-millimeter textures against the elastomer layer. Fig. \ref{fig:real result} shows the results for two objects. After crosstalk correction, input images exhibit improved channel separation. Our method reconstructs high-quality normals and detailed 3D geometry. The successful deployment and validation of our \textit{NearLightTactile} demonstrates the potential practical utility of the proposed approach. By mitigating crosstalk through targeted calibration, the method achieves normal reconstruction, 

% The input images, after crosstalk correction, show cleaner separation of the color channels. Our method successfully reconstructs high-quality surface normals and, through integration using a normal integration method\cite{Cao24} to obtain the corresponding mesh from the surface normal map, acquires detailed 3D geometry . The grooves, stripes, and bumps are all clearly visible and measurably accurate, demonstrating that our algorithm effectively translates from simulation to reality. Some minor artifacts persist in regions with extreme shadows or occlusions, which is an expected limitation given the single-image, three-light-source setup.
% The successful deployment of our algorithm on a physical sensor underlines its practical utility. We overcome the critical issue of crosstalk through a dedicated calibration step, enabling high-fidelity normal reconstruction of real-world objects. This validates our approach as a robust solution for applications requiring precise tactile perception, such as robotic manipulation and industrial inspection.

\vspace{-1.5ex}
\section{Conclusion}
\vspace{-1ex}

In this paper, we presented the first non-Lambertian CPS method under near-light conditions. By introducing a neural surface model to jointly constrain surface normals and depth, together with a neural BRDF model that enforces a mono-chromaticity constraint, our approach reduces reconstruction ambiguity and makes CPS tractable. The proposed framework achieves accurate shape recovery from a single multispectral snapshot and demonstrates practical feasibility through real-world validation with a compact visuo-tactile sensor.

\newpage

\vspace{-1.5ex}
\section{Acknowledgements}
\vspace{-1ex}
% This work was supported by Beijing-Tianjin-Hebei Basic Research Funding Program No. F2024502017, Beijing Municipal Science \& Technology Program No. Z251100007125021, Hebei Natural Science Foundation Project No. 242Q0101Z, National Natural Science Foundation of China (Grant No. 62472044, U24B20155, U23B2052.

% This work was supported by Beijing-Tianjin-Hebei Basic Research Funding Program No. F2024502017, Beijing Municipal Science \& Technology Program No. Z251100007125021, Hebei Natural Science Foundation Project No. 242Q0101Z, Science and Technology Commission of Shanghai Municipality No. 25ZR1401191, and the National Natural Science Foundation of China (Grant No. 62472044, U24B20155, U23B2052, and 52505029).

This work was supported by Beijing-Tianjin-Hebei Basic Research Funding Program No. F2024502017, the National Natural Science Foundation of China (Grant No. 62472044, U24B20155, U23B2052, and 52505029), Hebei Natural Science Foundation Project No. 242Q0101Z, Beijing Municipal Science \& Technology Program No. Z251100007125021, and Science and Technology Commission of Shanghai Municipality No. 25ZR1401191.

\bibliographystyle{IEEEbib}
\bibliography{strings,refs}

\end{document}